\documentclass{article}
\usepackage[margin=1.4in]{geometry}
\title{ZeRO: Memory Optimizations Toward Training Trillion Parameter Models}
\author{Samyam Rajbhandari$^*$, Jeff Rasley\footnote{Equal Contributors}, Olatunji Ruwase, Yuxiong He \\ \tt{\small \{samyamr, jerasley, olruwase, yuxhe\}@microsoft.com} }

\date{}
\usepackage{xspace}
\usepackage{graphicx}
\usepackage{multirow}
\usepackage{hhline}
\usepackage{comment}
\usepackage{adjustbox}
\usepackage{xcolor}
\usepackage{pifont}
\usepackage{hyperref}
\usepackage{subcaption}

\usepackage[switch,columnwise]{lineno} %TODO add back for SC review

\newcommand{\name}{\emph{ZeRO}\xspace}
\newcommand{\nameos}{\emph{ZeRO-OS}\xspace}

\begin{document}

% \linenumbers %TODO add back for SC review
%\pagenumbering{Roman}
\maketitle
\section*{Abstract}
Large deep learning models offer significant accuracy gains, but training billions to trillions of parameters is challenging. Existing solutions such as data and model parallelisms exhibit fundamental limitations to fit these models into limited device memory, while obtaining computation, communication and development efficiency.
We develop a novel solution, Zero Redundancy Optimizer (\name), to optimize memory, vastly improving training speed while increasing the model size that can be efficiently trained.  
%\name conquers the limitations of data- and model-parallelism while achieving the merits of both:
\name eliminates memory redundancies in data- and model-parallel training while retaining low communication volume and high computational granularity, allowing us to scale the model size proportional to the number of devices with sustained high efficiency. Our analysis on memory requirements and communication volume demonstrates: \name has the potential to scale beyond 1 \emph{Trillion} parameters using today's hardware.

We implement and evaluate ZeRO: it trains large models of over 100B parameter with super-linear speedup on 400 GPUs, achieving throughput of 15 Petaflops. This represents an 8x increase in model size and 10x increase in achievable performance over state-of-the-art.  
%\name-powered data parallelism simplifies large model training - scientists can train large models up to 12B parameters, 
In terms of usability, 
\name can train large models of up to 13B parameters (e.g., larger than Megatron GPT 8.3B and T5 11B) without requiring model parallelism which is harder for scientists to apply.
Last but not the least, researchers have used the system breakthroughs of \name to create the world's largest language model (17B parameters) with record breaking accuracy.

\section{Extended Introduction}
\label{sec:introduction}
Deep Learning (DL) models are becoming larger, and the increase in model size offers significant accuracy gain. In the area of Natural Language Processing (NLP), the transformers have paved way for large models like Bert-large (0.3B)~\cite{DBLP:journals/corr/bert}, GPT-2 (1.5B)~\cite{gpt-2}, Megatron-LM (8.3B)~\cite{megatronlm}, T5 (11B)~\cite{T5}. To enable the continuation of model size growth from 10s of billions to trillions of parameters, we experience the challenges of training them - they clearly do not fit within the memory of a single device, e.g., GPU or TPU, and simply adding more devices will not help scale the training.  

Basic data parallelism (DP) does not reduce memory per device, and runs out of memory for models with more than 1.4B parameters on current generation of GPUs with 32\,GB memory. Other existing solutions such as Pipeline Parallelism (PP), Model Parallelism (MP), CPU-Offloading, etc, make trade-offs between functionality, usability, as well as memory and compute/communication efficiency, but all of which are crucial to training with speed and scale. %(please see Sec.~\ref{sec:related-work} for more details). 

Among different existing solution for training large models, MP is perhaps the most promising. The largest models in the current literature, the 11B T5 model \cite{T5}, and Megatron-LM 8.3B \cite{megatronlm}, were both powered by model parallelism, implemented in Mesh-Tensorflow \cite{DBLP:journals/corr/mesh-tensor} and Megatron-LM\cite{megatronlm}, respectively. However, MP cannot scale much further beyond these models sizes. MP splits the model vertically, partitioning the computation and parameters in each layer across multiple devices, requiring significant communication between each layer. As a result, they work well within a single node where the inter-GPU communication bandwidth is high, but the efficiency degrades quickly beyond a single node \cite{megatronlm}. We tested a 40B parameter model using Megatron-LM across two DGX-2 nodes and observe about $5\,Tflops$ per V100 GPU (less than 5\% of hardware peak).    

So, how can we overcome the limitations of existing solutions and train large models more efficiently? To answer this question, we first analyze the full spectrum of memory consumption of the existing systems on model training and classify it into two parts:  1) For large models, the majority of the memory is occupied by \emph{model states} which include the optimizer states (such as momentum and variances in Adam~\cite{DBLP:journals/corr/Adam}), gradients, and  parameters. %(referred to as \emph{OGP} states in the paper). 
2) The remaining memory is consumed by activation, temporary buffers and unusable fragmented memory, which we refer to collectively as \emph{residual} states.
We develop \name --- Zero Redundancy Optimizer  --- to optimize memory efficiency on both while obtaining high compute and communication efficiency.
As these two parts face different challenges, we develop and discuss their solutions correspondingly.

{\bf Optimizing Model State Memory}
Model states often consume the largest amount of memory during training, but existing approaches such as DP and MP do not offer satisfying solution.  DP has good compute/communication efficiency but poor memory efficiency while MP can have poor compute/communication efficiency. More specifically, DP replicates the entire model states across all data parallel process resulting in redundant memory consumption;
while MP partition these states to obtain high memory efficiency, but often result in too fine-grained computation and expensive communication that is less scaling efficient.
Furthermore, all of these approaches maintain all the model states required over the entire
training process statically, even though not all model states are required all the time during
the training.
\begin{figure}[t!]
 \begin{center}
 \includegraphics[width=1.0\columnwidth]{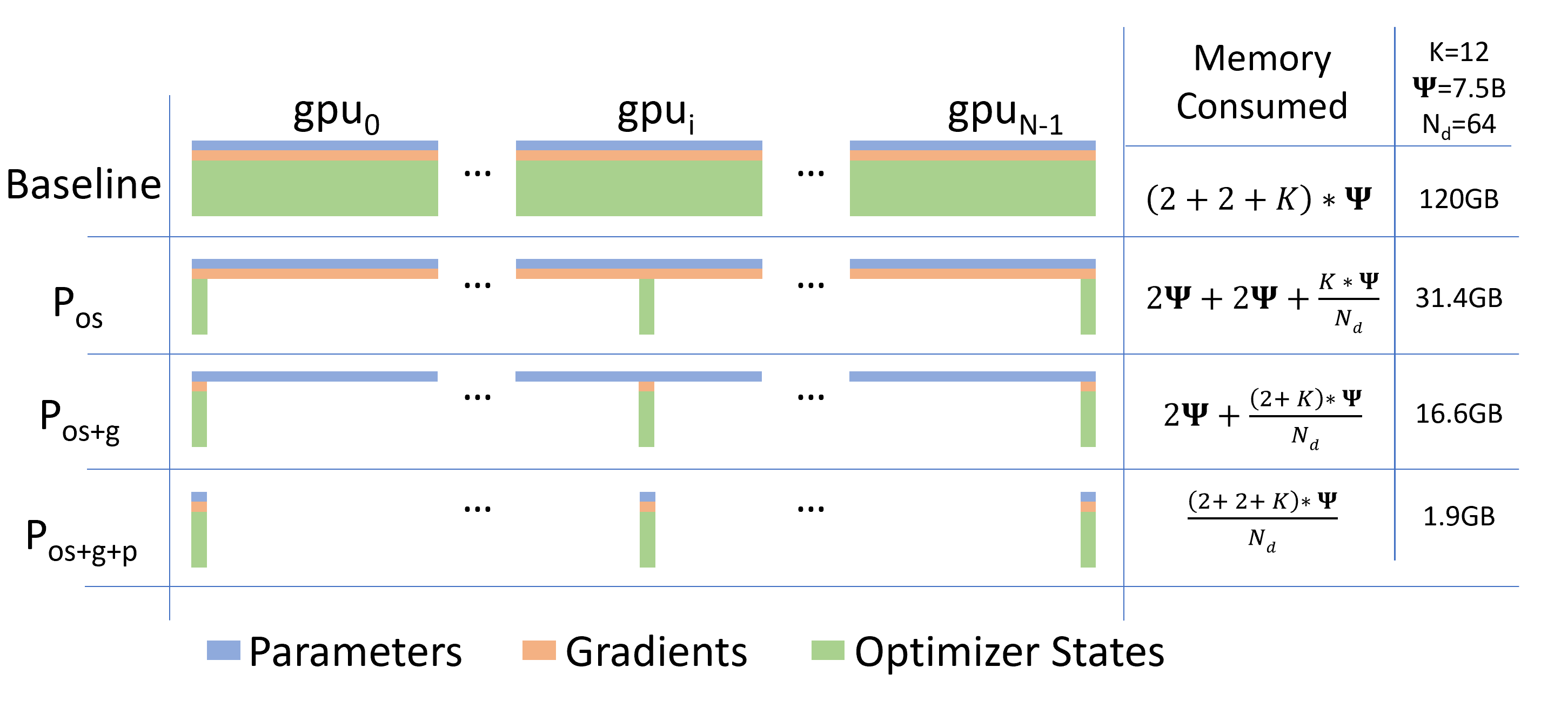}
 \caption{Comparing the per-device memory consumption of model states, with three stages of \name-DP optimizations. $\Psi$ denotes model size (number of parameters), $K$ denotes the memory multiplier of optimizer states, and $N_d$ denotes DP degree.  In the example, we assume a model size of $\Psi=7.5B$ and DP of $N_d=64$ with $K=12$ based on mixed-precision training with Adam optimizer. } 
 \label{fig:memory-consumption}
 \end{center}
 \end{figure}
Based on these observations, we develop \name-DP, ZeRO-powered data parallelism, that achieves the computation/communication efficiency of DP while achieving memory efficiency of MP.  \name-DP removes the memory state redundancies across data-parallel processes by \emph{partitioning} the model states instead of replicating them, and it retains the compute/communication efficiency by retaining the computational granularity and communication volume of DP using a dynamic communication schedule during training.  

\name-DP has three main optimization stages (as depicted in Figure \ref{fig:memory-consumption}), which correspond to the partitioning of optimizer states, gradients, and parameters. When enabled cumulatively:

1) Optimizer State Partitioning ($P_{os}$): 4x memory reduction, same communication volume as DP;

2) Add Gradient Partitioning ($P_{os+g}$): 8x memory reduction, same communication volume as DP; 

3) Add Parameter Partitioning ($P_{os+g+p}$): Memory reduction is linear with DP degree $N_d$. For example, splitting across 64 GPUs ($N_d$ = 64) will yield a 64x memory reduction. There is a modest 50\% increase in communication volume.

ZeRO-DP eliminates memory redundancies and makes the full aggregate memory capacity of a cluster available. With all three stages enabled, ZeRO can train a trillion-parameter model on just 1024 NVIDIA GPUs. A trillion-parameter model with an optimizer like Adam~\cite{DBLP:journals/corr/Adam} in 16-bit precision requires approximately 16 terabytes (TB) of memory to hold the optimizer states, gradients, and parameters. 16TB divided by 1024 is 16GB, which is well within a reasonable bound for a GPU (e.g., with 32GB of on-device memory).

%It reduces per-device memory footprint of a model \emph{linearly} with the increase in data parallelism degree while maintaining the communication volume close to that of the default data parallelism. \name-DP can fit models of \emph{arbitrary} size --- as long as there are sufficient number of devices to share the model states.  For example, our memory analysis shows that \name can fit a trillion parameter model on 1024 GPUs with data parallelism degree $N_d=1024$ (with more details in Section \ref{sec:summarymemoryoptimization}).

{\bf Optimizing Residual State Memory}
After \name-DP boosts memory efficiency for model states, the rest of the memory consumed by activations, temporary buffers, and unusable memory fragments could become a secondary memory bottleneck.  We develop \name-R to optimize the residual memory consumed by these three factors respectively.  

1) For activations (stored from forward pass in order to perform backward pass), we noticed checkpointing \cite{DBLP:journals/corr/ChenXZG16} helps but not sufficient for large models.  
Thus \name-R optimizes activation memory by identifying and removing activation replication in existing MP approaches through activation partitioning. It also offloads activations to CPU when appropriate.
%(see Sec.~\ref{sec:mp_activation_replication} for more details)

2) \name-R defines appropriate size for temporary buffers to strike for a balance of memory and computation efficiency. 

3) We observe fragmented memory during training due to variations in the lifetime of different tensors. Lack of contiguous memory due to fragmentation can cause memory allocation failure, even when enough free memory is available. \name-R proactively manages memory based on the different lifetime of tensors, preventing memory fragmentation.

%\name-R not only reduces memory usage but also improves training efficiency as we show in Sec.\ref{sec:evaluation}. 
\name-DP and \name-R combined together forms a powerful system of memory optimizations for DL training that we collectively refer to as \name.

\textbf{\name and MP}: Since \name eliminates the memory inefficiency in DP, it is natural to ask: Do we still need MP, and when?  How does \name work with MP?  With \name, MP becomes a less attractive option for the purpose of fitting large models alone.  \name-DP is at least as effective on reducing per-device memory footprint as MP, or more effective sometimes when MP cannot divide the model evenly. It also has comparable or better scaling efficiency. Furthermore, data parallelism is so easy to use that it is widely applicable across different workloads, while MP approaches today often need some work from model developers to revise their model, system developers to work out distributed operators, and existing work like Megatron-LM only supports a limited set of operators and models.

That being said, there are still cases where we want to leverage MP: i) When used with \name-R, MP can reduce activation memory footprint for very large models. ii) For smaller models where activation memory is not an issue, MP can also have benefits when aggregated batch size using DP alone is too big to have good convergence.\footnote{Prior work \cite{DBLP:journals/corr/batch-scaling} shows, very large batch size could slow down convergence.  For given model and data, there is a measure of critical-batch size, where increasing batch size further slows down convergence.  The detailed discussion of this topic is beyond the scope of the paper.}  In those case, one can combine \name with MP to fit the model with an acceptable aggregated batch size.

We show that \name can be combined with MP, resulting in a max theoretical memory reduction of $N_d \times N_m$ times on each device with a DP degree of $N_d$ and MP degree of $N_m$. 
This could allow us to fit a trillion parameter model on 1024 GPUs with 16-way model parallelism (within each DGX2 node) and 64-way data parallelism across nodes, and run it efficiently using a modest batch size!

\begin{figure}[t!]
 \begin{center}
 \includegraphics[width=1.0\columnwidth]{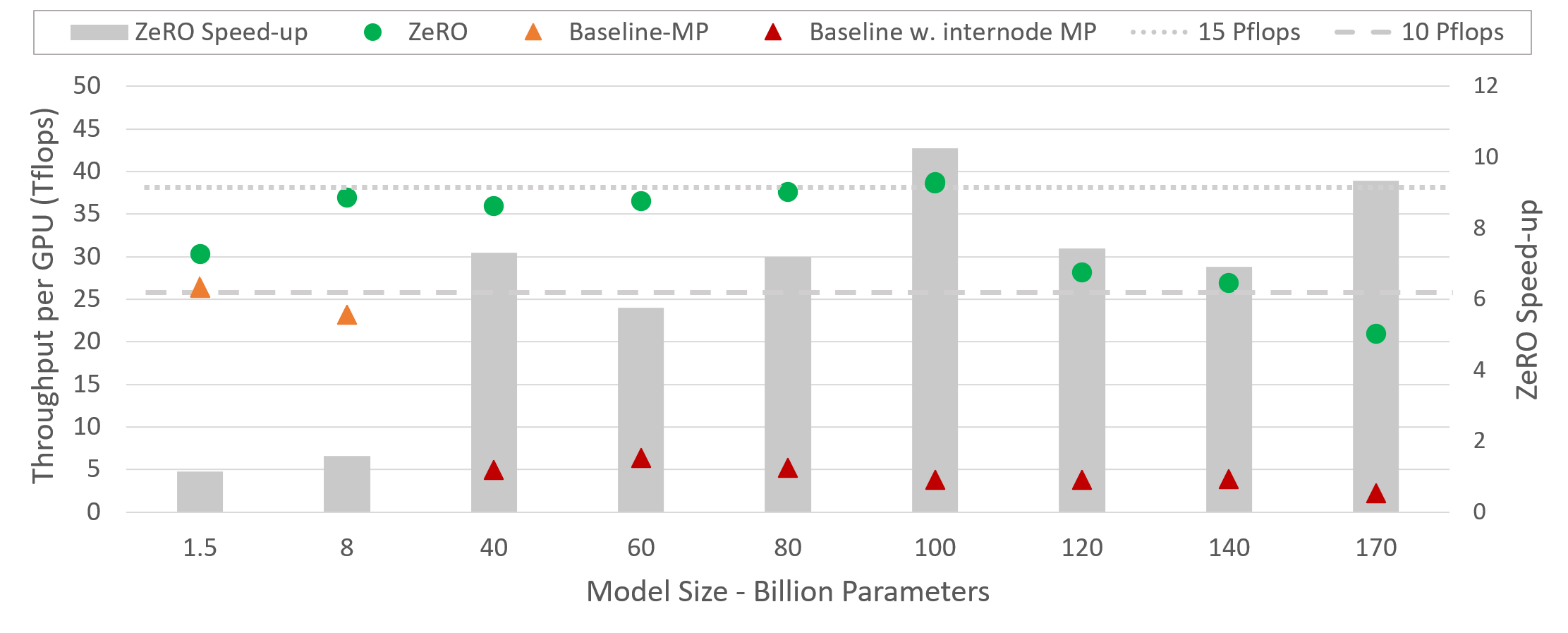}
 \caption{\name training throughput and speedup w.r.t SOTA baseline for varying model sizes.  For \name, the MP always fit in a node, while for baseline, models larger than 40B require MP across nodes.} 
 \label{fig:billion_parameter_speedup}
 \end{center}
 \end{figure}

\begin{figure}[t!]
 \begin{center}
 \includegraphics[width=1.0\columnwidth]{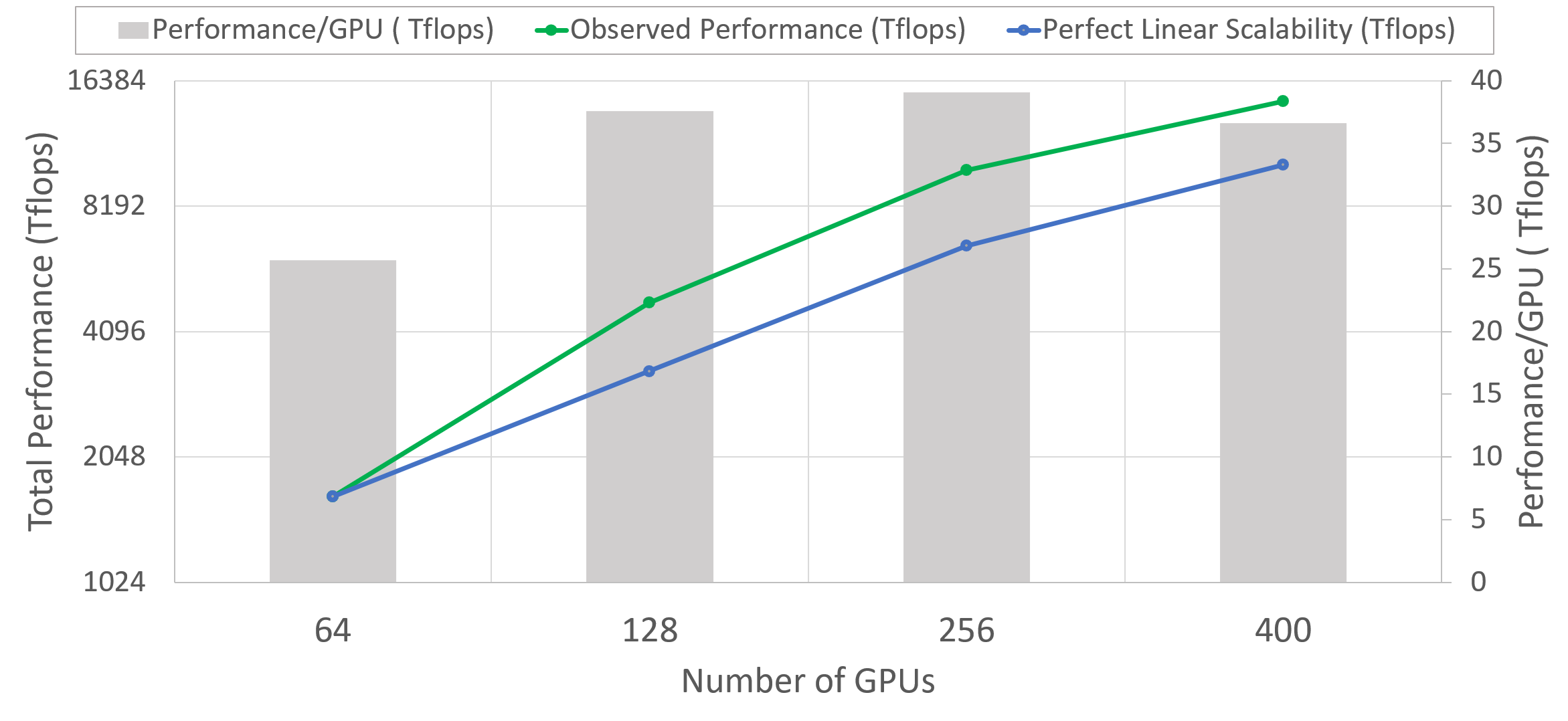}
 \caption{Superlinear scalability and per GPU training throughput of a 60B parameter model using \name-100B.} 
 \label{fig:hyperscale_60B}
 \end{center}
 \end{figure}

% \begin{figure*}[t!]
%     \centering
%     \begin{minipage}{.55\textwidth}
%         \centering
%         \includegraphics[width=\linewidth]{model_size_and_speedup.PNG}
%         \caption{\name training throughput and speedup w.r.t SOTA baseline for varying model sizes.  For \name, the MP always fit in a node, while for baseline, models larger than 20B require MP across nodes. }\label{fig:billion_parameter_speedup}
%     \end{minipage}
%     \quad
%     \begin{minipage}{.4\textwidth}
%         \centering
%         \includegraphics[width=\linewidth]{hyperscale_60B_model.PNG}
%         \caption{Superlinear scalability and per GPU training throughput of a 60B parameter model using \name-100B.}\label{fig:hyperscale_60B}
%     \end{minipage}
% \end{figure*}

{\bf Implementation \& Evaluation}
The complete set of optimizations in \name could allow us to run models with trillion parameters on the high-end hardware cluster today (e.g., with 1K V100 GPUs), however, the hardware compute capacity is still too limited and training time can be impractically long ($>$1 year).  Therefore, our focus for this implementation is to efficiently support models with 10x parameters ($\sim$100B parameters) than state-of-the-art (SOTA) while still being within reach of the compute capabilities of current hardware. We implement and evaluate a subset of optimizations in \name called \name-100B --- $P_{os+g}$ of \name-DP plus ZeRO-R --- that allow us to achieve this goal. The results show:

\underline{Model Size} Combined with MP, \name-100B runs 170B parameter models efficiently, while the existing system like using Megatron alone cannot scale efficiently beyond 40B parameters, as shown in Figure~\ref{fig:billion_parameter_speedup}.
This is an over 8x increase in model size compared to SOTA.

\underline{Speed} Improved memory efficiency powers higher throughput and faster training. 
As shown in Figure~\ref{fig:billion_parameter_speedup}, \name runs 100B parameter models on a 400 Nvidia V100 GPU cluster with over 38 TFlops per GPU, and aggregate performance over 15 Petaflops. This is more than 10x improvement in training speed compared to SOTA for the same model size.

\underline{Scalability} We observe super linear speedup in the regime of 64-400 GPUs, where the performance more than doubles when we double the number of GPUs. This is a property of \name-DP which reduces the memory footprint of the model states as we increase the DP degree allowing us to fit larger batch sizes per GPU resulting in better performance. We expect this behaviour to continue further as we increase the number of GPUs beyond 400.

\underline{Democratization of Large Model Training} \name-100B powers data scientist to train models with up to 13B parameters without any MP or PP that requires model refactoring, where 13B is more parameters than the largest model in literature (T5 with 11B parameters). Data scientists can thus experiment freely with large models without worrying about parallelism. In comparison, exist systems (e.g., PyTorch Distributed Data Parallel) runs out of memory with 1.4B parameter models. 
%Furthermore, MP requires high-bandwidth interconnect such as NVLINK/NVSwitch; \name-100B allows them to be trained more efficiently on clusters that don't have these high-end intra-node interconnect.

\underline{New SOTA Model} \name powers the largest language model with 17B parameters and record-breaking accuracy, Turing-NLG~\cite{t-nlg}.
We share \name as a part of our open source DL training optimization library called DeepSpeed\footnote{https://github.com/microsoft/deepspeed}.
We plan to release all implementations described in this paper by end of May 2020 and extend it further to support 1 trillion parameters by enabling \name-DP stage 3 partitioning parameters ($P_{os+g+p}$). We plan to make \name fully accessible to the DL community to catalyze the evolution and democratization of large model training at scale.  

\section{Related Work}
\label{sec:related-work}
\subsection{Data, Model and Pipeline Parallelism}
Parallelization is a key strategy on training large models at scale. For a model that fits in the device memory for training, data parallelism (DP) is used to scale training to multiple devices. In DP, model parameters are replicated on each device. At each step, a mini-batch is divided evenly across all the data parallel processes, such that each process executes the forward and backward propagation on a different subset of data samples, and uses averaged gradients across processes to update the model locally.  

When a model does not fit in the device memory, model parallelism (MP) \cite{DBLP:journals/corr/mesh-tensor, megatronlm} and pipeline parallelism (PP) \cite{GPipe, DBLP:journals/corr/pipedream} split the model among processes, in vertical and horizontal way respectively. Sec.~\ref{sec:introduction} discussed how \name relates to DP and MP. We now discuss PP and how it relates to reducing memory consumption.

PP splits a model horizontally across layers running each partition on a different device and use micro-batching to hide the pipeline bubble \cite{GPipe,DBLP:journals/corr/pipedream}. Model functionalities such as tied-weights and batch-normalization are difficult to implement due to horizontal splitting and micro-batching, respectively. Popular PP implementation such as G-pipe \cite{GPipe} partitions both model parameters and total activations but requires a batch size proportional to number of pipeline partitions to hide the pipeline bubble. The large batch size can affect the convergence rate, while also requiring significant memory to store activations. A different implementation of PP in PipeDream \cite{narayanan2019pipedream} keeps multiple copies of stale parameters to hide the pipeline bubble without increasing the batch size significantly, making it less memory efficient. Additionally, the implementation is not equivalent to the standard DL training and has implications on training convergence. 
\begin{comment}
Similar to MP, PP is not easily accessible to model scientists: It either requires them to re-write their model using a PP framework, or it requires complex static and run-time analysis support to split a model graph into load balanced pipeline stages.  This requires altering the core of deep learning framework itself (e.g., PyTorch, TensorFlow) \cite{DBLP:journals/corr/pipedream}.  {\color{red} [Not sure if the later is a problem for model scientist]} Furthermore, PP requires a significant increase in batch size to hide the pipeline bubble overhead compared to DP.  Large batch size can lead to slower convergence as discussed in many studies \cite{}.  Alternately, PP requires keeping multiple model states \cite{DBLP:journals/corr/pipedream} to avoid the increase in batch size which not only increases memory but also has convergence implications.  Furthermore, activation memory consumption in PP is inherently imbalanced, as the pipeline stages between forward and backward propagation vary depending on the pipeline stage in between, i.e., initial stages require more memory than the latter stages since the lifetime of activations in the initial stage are longer. 
\end{comment}
In contrast, \name obtains the same or better memory efficiency than PP without incurring functionality, performance and convergence related restrictions of PP.

\begin{comment}
\name addresses the limitations of DP, MP and PP.  \name can scale across multiple nodes without impact on efficiency, while MP can only scale efficiently within a node. \name does not have any load balancing issues or bubble overhead found in PP. Additionally, \name does not require any changes to the model making it as easy to use as pure DP.
\end{comment}

\subsection{Non-parallelism based approach to reduce memory}
In addition to MP and PP, there are multiple lines of work that target reducing memory overheads of DL training. 
\subsubsection{Reducing Activation Memory}
Multiple efforts have focused on reducing the memory footprint of activations through compression~\cite{jain2018gist}, activation checkpointing~\cite{DBLP:journals/corr/ChenXZG16, Jain2019CheckmateBT}, or live analysis \cite{DBLP:journals/corr/abs-1801-04380}. These efforts are complimentary and can work together with \name. In fact, activation memory reduction in \name-R works in parallel with activation checkpointing. 
\subsubsection{CPU Offload}
\cite{layer2layer, 7783721} exploit heterogeneous nature of today's compute nodes, offloading model states to CPU memory through algorithmic design or virtualized memory, respectively. Up to $50\%$ of training time can be spent on GPU-CPU-GPU transfers \cite{layer2layer}. \name differs in that it reduces the memory consumption significantly without storing the model states to CPU memory whose bandwidth is severely constrained due to PCI-E. On rare cases, \name-R may offload just the activation checkpoints for very large models to improve performance (see Sec.~\ref{sec:p_a} for details). 
\subsubsection{Memory Efficient Optimizer}
\cite{DBLP:journals/corr/adafactor,Anil2019MemoryEfficientAO} focus on reducing memory consumption of adaptive optimization methods by maintaining coarser-grained statistics of model parameters and gradients, with potential impact on model convergence guarantees. \name is orthogonal to these efforts, and its optimizations 
do not change the model optimization method or affect model convergence, but effectively reduce memory footprint of optimizer states and gradients per device.
\subsection{Training Optimizers }
Adaptive optimization methods~\cite{10.5555/Adagrad,DBLP:journals/corr/Adam,DBLP:journals/corr/You-LARS,DBLP:journals/corr/You-LAMB} are crucial to achieving SOTA performance and accuracy for effective model training of large models.  Compared to SGD, by maintaining fine-grained first-order and second-order statistics for each model parameter and gradient at the cost of significant memory footprint. \name can reduce the memory footprint of these optimizers by orders of magnitude, making these sophisticated optimization methods practical for training large models on hardware with modest device memory. It also makes it possible to develop and use even more complex and memory hungry optimizers that may have better convergence.

\section{Where Did All the Memory Go?}
Let's take a step back to examine the memory consumption of the current training system.  For example,  a 1.5B parameter GPT-2 model requires 3GB of memory for its weights (or parameters) in 16-bit precision, yet, it cannot be trained on a single GPU with 32GB memory using Tensorflow or PyTorch.  One may wonder where all the memory goes.
During model training, most of the memory is consumed by {\it model states}, i.e., tensors comprising of pptimizer states, gradients, and parameters. Besides these model states, the rest of the memory is consumed by activations, temporary buffers and fragmented memory which we call \emph{residual states}. We look at the memory consumption from both in details.
\begin{comment}
and iii) temporary buffers. It is possible to trivially  Here we look at the memory consumed by latter two of the three.
\end{comment}

\subsection{Model States: Optimizer States, Gradients and Parameters} Majority of the device memory is consumed by model states during training.  Consider for instance, Adam~\cite{DBLP:journals/corr/Adam}, one of the most popular optimizers for DL training. Adam requires storing two optimizer states, i) the time averaged momentum and ii) variance of the gradients to compute the updates. Therefore, to train a model with ADAM, there has to be enough memory to hold a copy of both the momentum and variance of the gradients. In addition, there needs to be enough memory to store the gradients and the weights themselves. Of these three types of the parameter-related tensors, the optimizer states usually consume the most memory, specially when mixed-precision training is applied.

\textbf{Mixed-Precision Training} The state-of-the-art approach to train large models on the current generation of NVIDIA GPUs is via mixed precision (fp16/32) training~\cite{micikevicius2017mixed}, where parameters and activations are stored as fp16, enabling the use of the high throughput tensor core units~\cite{nvidia-volta-arch} on these GPUs. During mixed-precision training, both the forward and backward propagation are performed using fp16 weights and activations. However, to effectively compute and apply the updates at the end of the backward propagation, the mixed-precision optimizer keeps an fp32 copy of the parameters as well as an fp32 copy of all the other optimizer states. 

Let's take Adam as a concrete example.  Mixed precision training of a model with $\Psi$ parameters using Adam requires enough memory to hold an $fp16$ copy of the parameters and the gradients, with memory requirements of $2\Psi$ and $2\Psi$ bytes respectively.  In addition, it needs to hold the optimizer states: an $fp32$ copy of the parameters, momentum and variance, with memory requirements of $4\Psi$, $4\Psi$, and $4\Psi$ bytes, respectively.  Let's use $K$ to denote the memory multiplier of the optimizer states, i.e., the additional memory required to store them is $K\Psi$ bytes. Mixed-precision Adam has $K=12$. In total, this results in $2\Psi + 2\Psi + K\Psi = 16\Psi$ bytes of memory requirement. 
For a model such as GPT-2 with $1.5$ Billion parameters, this leads to a memory requirement of at least $24\,GB$, which is significantly higher than the meager $3\,GB$ of memory required to hold the $fp16$ parameters alone.

\subsection{Residual Memory Consumption}
\textbf{Activations} can take up a significant amount of memory \cite{DBLP:journals/corr/ChenXZG16} during training.  As a concrete example, the 1.5B parameter GPT-2 model trained with sequence length of 1K and batch size of 32 requires about 60\,GB of memory\footnote{The activation memory of a transformer-based model is proportional to the number of transformer layers $\times$ hidden dimensions $\times$  sequence length $\times$ batch size.  For a GPT-2 like architecture the total activations is about  $12 \times hidden\_dim \times batch \times seq\_length \times transformer\_layers$.}.  Activation checkpointing (or activation recomputation) is a common approach to reduce the activation memory by approximately the square root of the total activations at the expense of $33\%$ re-computation overhead \cite{DBLP:journals/corr/ChenXZG16}. This would reduce the activation memory consumption of this model to about 8\,GB.

Despite the significant reduction, the activation memory can grow quite large for bigger models even with activation checkpointing. For example, a GPT-like model with 100 billion parameters requires around 60 GB of memory for batch size 32, even when using activation checkpointing.

\begin{comment}
the amount of activation memory for a transformer based language model is proportional to the batch size, sequence length, hidden dimension and the number of layers . 
Activations take a non-trivial amount of memory even when using techniques such as activation checkpointing which significantly reduce the memory required by the activations at the expense of a $33\%$ re-computation overhead\cite{}.

For example, lets consider the 1.5B parameter GPT-2 model. Without activation checkpointing a transformer model produces $O(\frac{model\_size}{hidden\_dim}*seq\_length*batch$) activation, while with activation checkpointing it produces $O(\frac{model\_size}{hidden\_dim}*seq\_length*batch$) activations if we stored the activations only at the transformer layer boundary. For a batch size of 32, the GPT-2 models will therefore produce about 60GB of activations in fp16, while with activation checkpointing, it produces about 5 GB of activations.
\end{comment}

\textbf{Temporary buffers} used for storing intermediate results consumes non-trivial amount of memory for large models.   Operations such as gradient all-reduce, or gradient norm computation tend to fuse all the gradients into a single flattened buffer before applying the operation in an effort to improve throughput. For example, the bandwidth of all-reduce across devices improves with large message sizes. While the gradient themselves are usually stored as fp16 tensors, the fused buffer can be an fp32 tensor depending on the operation. When the size of the model is large, these temporary buffer sizes are non-trivial. For example, for a model with 1.5B parameters, a flattened fp32 buffer would required $6\,GB$ of memory. 

\textbf{Memory Fragmentation: }  So far we have discussed the actual memory consumption during training. Additionally, it is possible to run out of usable memory even when there is plenty of available memory. This can happen with memory fragmentation. A request for a memory will fail if there isn't enough contiguous memory to satisfy it, even if the total available memory is larger than requested. We observe significant memory fragmentation when training very large models, resulting in out of memory issue with over 30\% of memory still available in some extreme cases.
\begin{comment}

Training a large model with many layers can runs out of memory during training, even when there is plenty of available memory. 

We find that activation checkpointing and backward propagation can cause heavy memory fragmentation, making significant portion of the GPU memory unusable, even when there is plenty of available memory. 
The memory fragmentation is caused by activation checkpointing and backward propagation allocating long term and shot term memory in an interleaved fashion. During the forward propagation, temporary buffers are created in between the creation of each activation checkpoint, causing memory fragmentation. Similarly, in backward propagation working memory is allocated between computation of the gradients, also causing memory fragmentation. We find that for models with hundreds of layers, memory fragmentation can make over 30\% of the GPU memory unusable.
\end{comment}

\section{\name: Insights and Overview}
%[Transition: the scaling limitation of MP and memory inefficiency of DP motivates \name: Can we overcome the memory inefficiency of DP but still] 
 \name has two sets of optimizations: i) \name-DP aimed at reducing the memory footprint of the model states, and ii) \name-R targeted towards reducing the residual memory consumption.  We present an overview of the optimizations and the insights behind, which allows \name to reduce memory footprint while remaining efficient. Please note efficiency is a key here: without this constraint, trivial solutions like moving all the parameter states to the CPU memory, or  increasing the MP degree arbitrarily can reduce memory footprint.   
\begin{comment}Now that we know where all the memory goes, how can we reduce the memory footprint and use it effectively without sacrificing efficiency?  Our solution, for reducing memory footprint without sacrificing efficiency is based on three sets of insights. 
\end{comment}

\subsection{Insights and Overview: \name-DP}
\name powered DP is based on three key insights:

{\it a)} DP has better scaling efficiency than MP because MP reduces the granularity of the computation while also increasing the communication overhead. Beyond a certain point, lower computational granularity reduces the efficiency per GPU, while the increased communication overhead, hiders the scalability across GPUs, especially when crossing node boundaries. On the contrary, DP has both higher computational granularity and lower communication volume, allowing for much higher efficiency.

{\it b)} DP is memory inefficient as model states are stored redundantly across all data-parallel processes. On the contrary, MP partitions the model states to obtain memory efficiency.

{\it c)} Both DP and MP keep all the model states needed over the entire training process, but not everything is required all the time.  For example, parameters corresponding to each layer is only needed during the forward propagation and backward propagation of the layer. %At all other times, these parameters are not needed.  

Based on these insights, \name-DP retains the training efficiency of DP while achieving the memory efficiency of MP. \name-DP \emph{partitions} the model states instead of replicating them (Section \ref{sec:memoryoptimization}) and uses a dynamic communication schedule that exploits the intrinsically temporal nature of the model states while minimizing the communication volume (Section \ref{sec:communication}). By doing so, \name-DP reduces per-device memory footprint of a model \emph{linearly} with the increased DP degree while maintaining the communication volume close to that of the default DP, retaining the efficiency.
\subsection{Insights and Overview: \name-R}
\label{sec:mp_activation_replication}
\subsubsection{Reducing Activation Memory}
%Memory optimizations for reducing the memory footprint for activation states is based on two key insights:
Two key insights are:

{\it a)} MP partitions the model states but often requires replication of the activation memory. For example, if we split the parameters of a linear layer vertically and compute them in parallel across two GPUs, each GPU requires the entire activation to compute its partition

{\it b)} For models such as GPT-2 or larger, the arithmetic intensity (ratio of the amount of computation per iteration to amount of activation checkpoints per iteration) is very large ($\ge 10K$) and increases linearly with hidden dimension making it possible to hide the data-movement cost for the activation checkpoints, even when the bandwidth is low.

\name removes the memory redundancies in MP by \emph{partitioning} the activations checkpoints across GPUs, and uses allgather to reconstruct them on demand. The activation memory footprint is reduced proportional to the MP degree. For very large models, \name can even choose to move the activation partitions to the CPU memory, while still achieving good efficiency due to large arithmetic intensity in these models. 
\subsubsection{Managing Temporary buffers}
\name-R uses constant size buffers to avoid temporary buffers from blowing up as the model size increases, while making them large enough to remain efficient. 

\subsubsection{Managing fragmented Memory} Memory fragmentation is a result of interleaving between short lived and long lived memory objects. During the forward propagation activation checkpoints are long lived but the activations that recomputed are short lived. Similarly, the backward computation, the activation gradients are short lived while the parameter gradients are long lived. Based on this insight, \name performs on-the-fly memory defragmentation by moving activation checkpoints and gradients to pre-allocated contiguous memory buffers. This not only increases memory availability but also improves efficiency by reducing the time it takes for the memory allocator to find free contiguous memory.  

\section{Deep Dive into \name-DP}\label{sec:memoryoptimization}
%\name primarily reduces the training memory consumption by removing memory redundancies across data parallel process. As described in Sec. \ref{data-parallel}, each data parallel process keeps a copy of the optimizer states and the parameters, and uses the averaged gradients to compute the updates to its copy of the parameters. Therefore, the optimizer states, averaged gradients and the parameters themselves are redundant across the data parallel processes. \name partitions these aforementioned tenors across data parallel process to reduce the memory requirements by eliminating the redundancy. \name further optimizes the memory consumption via communication and computation re-scheduling such that the size of the temporary buffers required for these operations is significantly reduced.
While the existing DP approach replicates the model states at each device and introduces significant memory overhead, \name-DP eliminates this memory redundancy by partitioning them --- optimizer states, gradients and parameters --- across data parallel processes.  
Figure \ref{fig:memory-consumption} quantifies and visualizes the memory requirement with and without \name-DP. The figure shows the memory footprint after partitioning (1) optimizer state, (2) gradient and (3) parameter redundancies accumulatively.  We refer to them as the three optimization phases of \name-DP: $P_{os}$, $P_g$, and $P_p$, which we elaborate below. 
\begin{comment}
Furthermore, \name removes large buffers proportional to model size with performance-efficient constant-size buffers (Section \ref{sec:buffer}) to address the memory overhead of large temporary buffers.  The memory optimizations require careful computation, communication rescheduling and mapping across data parallel process. 
Next we go through each of these techniques in details.
\end{comment}
\subsection{P$_{os}$ : Optimizer State Partitioning }\label{sec:pos}
For a DP degree of $N_d$, we group the optimizer states into $N_d$ equal partitions, such that the $i^{th}$ data parallel process only updates the optimizer states corresponding to the $i^{th}$ partition. Thus, each data parallel process only needs to store and update $\frac{1}{N_d}$ of the total optimizer states and then only update $\frac{1}{N_d}$ of the parameters.
%Note that here we only eliminate the optimizer state redundancy, but not the parameter redundancy. Since each process only has $\frac{1}{N_d}$ of the overall optimizer state, it can only update $\frac{1}{N_d}$ of the parameters. 
We perform an all-gather across the data parallel process at the end of each training step to get the fully updated parameters across all data parallel process. %We will refer to this optimization as $P_{os}$.

\textbf{Memory Savings: } As shown in Figure 1, the memory consumption after optimizing state partition reduces from $4\Psi+K\Psi$ to $4\Psi + \frac{K\Psi}{N_d}$.  As the concrete example depicted in Figure \ref{fig:memory-consumption}, a 7.5 B parameter model requires 31.4GB of memory using $P_{os}$ with 64-way DP ($N_d = 64$), while requiring 120 GB with standard DP.
%lets take the 1.5B GPT-2 model training on 8 GPUs ($N_d = 8$) using mixed precision training with the ADAM optimizer. As discussed in Sec.~\ref{memory-requirements}, this requires at least $ 4M  + 12M = 6 + 18 = 24\ gb$ of memory to store the parameters, gradients and the optimizer states. By eliminating optimizer state redundancy, the memory requirement reduces to $6+\frac{18}{N_d} = 8.25\ gb$. 
Furthermore, when $N_d$ is large, the memory requirement on model states reduces from $4\Psi+12\Psi=16\Psi$ bytes to $ 4 \Psi + \frac{12 \Psi}{N_d} \approx 4 \Psi$ bytes, leading to a 4x reduction. 

\subsection{P$_g$: Gradient Partitioning}\label{sec:pg}
As each data parallel process only updates its corresponding parameter partition, it only needs the reduced gradients for the corresponding parameters. Therefore, as each gradient of each layer becomes available during the backward propagation, we only reduce them on the data parallel process responsible for updating the corresponding parameters. After the reduction we no longer need the gradients and their memory can be released. This reduces the memory footprint required to hold the gradients from $2\Psi$ bytes to $\frac{2\Psi}{N_d}$. 

Effectively this is a Reduce-Scatter operation, where gradients corresponding to different parameters are reduced to different process. To make this more efficient in practice, we use a bucketization strategy, where we bucketize all the gradients corresponding to a particular partition, and perform reduction on the entire bucket at once. 
This is similar in spirit to how NVIDIA's AMP \cite{nvidia-apex} optimizer bucketizes the all-reduce gradient computation to overlap communication and computation. In our case we perform a reduction instead of an all-reduce at the partition boundaries to reduce memory footprint and overlap computation and communication. %We refer to this optimization as P$_g$.

\textbf{Memory Savings:} By removing both gradient and optimizer state redundancy, we reduce the memory footprint further down to $ 2 \Psi + \frac{14\Psi}{N_d} \approx 2 \Psi$. 
%In case of GPT-2, this reduces the memory footprint down to meager $3 gb$ to store a single copy of the parameters in fp16.
As the example in Figure 1, a 7.5 B parameter model requires only 16.6 GB of memory using $P_{os+g}$ with 64-way DP ($N_d = 64$), while requiring 120 GB with standard DP.
When $N_d$ is large,  the memory requirement of model states reduces from $2\Psi+14\Psi=16\Psi$ bytes to $ 2 \Psi + \frac{14 \Psi}{N_d} \approx 2 \Psi$ bytes, leading to a 8x reduction.
  \begin{table*}
        \centering
        %\resizebox{2*\columnwidth}{!}{
        \begin{tabular}{|c||c|c|c||c|c|c||c|c|c|}
        \hline
        \multicolumn{1}{|c}{\multirow{2}{*}{DP}}&\multicolumn{3}{||c}{\textbf{7.5B Model} (GB)} &\multicolumn{3}{||c}{\textbf{128B Model} (GB)} & \multicolumn{3}{||c|}{\textbf {1T Model} (GB)} \\
        \hhline{~---------}
        &P$_{os}$&P$_{os+g}$&P$_{os+g+p}$&P$_{os}$&P$_{os+g}$&P$_{os+g+p}$&P$_{os}$&P$_{os+g}$&P$_{os+g+p}$\\
        \hline
        1&120&120&120&2048&2048&2048&16000&16000&16000\\
        4&52.5&41.3&\textbf{30}&896&704&512&7000&5500&4000\\
        16&35.6&\textbf{21.6}&7.5&608&368&128&4750&2875&1000\\
        64&\textbf{31.4}&16.6&1.88&536&284&\textbf{32}&4187&2218&250\\
        256&30.4&15.4&0.47&518&263&8&4046&2054&62.5\\
        1024&30.1&15.1&0.12&513&257&2&4011&2013&\textbf{15.6}\\
        \hline
        \end{tabular}
        %}
     \caption{Per-device memory consumption of different optimizations in \name-DP as a function of DP degree . Bold-faced text are the combinations for which the model can fit into a cluster of 32GB V100 GPUs.}
     \label{tab:memory-consumption}
 \end{table*}

\subsection{P$_p$: Parameter Partitioning}\label{sec:pp}
Just as with the optimizer states, and the gradients, each process only stores the parameters corresponding to its partition. When the parameters outside of its partition are required for forward and backward propagation, they are received from the appropriate data parallel process through broadcast. While this may seem to incur significant communication overhead at first glance, we show that this approach only increases the total communication volume of a baseline DP system to $1.5$x, while enabling memory reduction proportional to $N_d$.

\textbf{Memory Savings:} With parameter partitioning, we reduce the memory consumption of an $\Psi$ parameter model from $16\Psi$ to $\frac{16\Psi}{N_d}$.  As the example in Figure 1, a 7.5 B parameter model requires 1.9 GB  of model-state memory using $P_{os+p+g}$ with 64-way DP ($N_d = 64$), while requiring 120 GB with standard DP.  This has a profound implication: \emph{\name powers DP to fit models with arbitrary size— as long as there are sufficient number of devices to share the model states}. \begin{comment}
For example, on 32 GPU cluster with 32-way DP, we can run a 64B parameter model per GPU without any MP, while on a cluster of 512 GPU with 16-way MP and 32-way DP, we can train a $64 * 16B = 1T$ parameter model.  {\color{green} [Same example as in Figure 1 please]}
\end{comment}
\subsection{Implication on Model Size}\label{sec:summarymemoryoptimization}
The three phases of partitioning $P_{os}$, $P_{os+g}$, and $P_{os+g+p}$ reduces the memory consumption of each data parallel process on model states by up to 4x, 8x, and $N_d$ respectively.  
%$C_b$ allows us to use constant-size buffers instead of growing buffer size linearly with the increase of model size.
  Table \ref{tab:memory-consumption} analyzes model-state memory consumption of a few example models under the 3 stages of \name-DP optimizations for varying DP degree.  Without \name, the memory consumption is equal to the first row in the table, regardless of the DP degree. Note that, with $N_d=64$, \name can train models with up to 7.5B, 14B, and 128B parameters using $P_{os}$, $P_{os+g}$, and $P_{os+g+p}$, respectively.  When $N_d=1024$, \name with all of its optimizations enabled ($P_{os+g+p}$) could train models with 1 {\sc{Trillion}} parameters!  Or potentially, models with {\sc{Arbitrary}} size!  Without \name, the largest model DP alone can run has less than 1.5 Billion parameters.

\section{Deep Dive into \name-R}
\subsection{$P_a$: Partitioned Activation Checkpointing}
\label{sec:p_a}
As discussed in \ref{sec:mp_activation_replication}, MP by design requires a replication of the activations, resulting in redundant copies of the activations across model parallel GPUs. \name eliminates this redundancy by partitioning the activations, and only materializes them in a replicated form one activation layer at a time, right before the activation is used in computation. More specifically, once the forward propagation for a layer of a model is computed, the input activations are partitioned across all the model parallel process, until it is needed again during the backprogation. At this point, \name uses an all-gather operation to re-materialize a replicated copy of the activations. We refer to this optimization as $P_a$. It works in conjunction with activation checkpointing \cite{DBLP:journals/corr/ChenXZG16}, storing partitioned activation checkpoints only instead of replicated copies. Furthermore, in the case of very large models and very limited device memory, these partitioned activation checkpoints can also be offloaded to the CPU reducing the activation memory overhead to nearly zero at an additional communication cost, which we will discuss in \ref{sec:communication}. We refer to this as $P_{a+cpu}$. 

\textbf{Memory Saving} With partitioned activation checkpointing, \name reduces the activation footprint by a factor proportional to the MP degree. Consider training a 100B model shown in Table~\ref{tab:model-configuration} with a batch size of 32, sequence length of 1024 and a MP degree of 16. If we checkpoint a single activation for each transformer layer, it would require about 33 GB of memory per GPU just to store the activation checkpoints. But with P$_a$ in \name, it can be reduced to about 2 GB per GPU. Furthermore, this 2GB can be offloaded  to the CPU reducing the memory footprint for activations to nearly zero.
\subsection{C$_B$: Constant Size Buffers}\label{sec:buffer}
\name carefully selects the sizes of the temporal-data buffers to balance memory and compute efficiency.
During training, the computational efficiency of some operations can be highly dependent on the input size, with larger inputs achieving higher efficiency. For example, a large all-reduce operation achieves much higher bandwidth than a smaller one. Hence, to get better efficiency, high performance libraries such as NVIDIA Apex or Megatron fuses all the parameters into a single buffer before applying these operations. However, the memory overhead of the fused buffers is proportional to the model size, and can become inhibiting. For example, for a 3B parameter model, a 32-bit fused buffer will require 12 GB of memory. To address this issue, we simply use a performance-efficient constant-size fused buffer when the model becomes too large. By doing so, the buffer size does not depend on the model size, and by keeping the buffer size large enough, we can still achieve good efficiency.

\subsection{M$_D$: Memory Defragmentation}
Memory fragmentation in model training occurs as a result of activation checkpointing and gradient computation. During the forward propagation with activation checkpointing, only selected activations are stored for back propagation while most activations are discarded as they can be recomputed again during the back propagation. This creates an interleaving of short lived memory (discarded activations) and long lived memory (checkpointed activation), leading to memory fragmentation. Similarly, during the backward propagation, the parameter gradients are long lived, while activation gradients and any other buffers required to compute the parameter gradients are short lived. Once again, this interleaving of short term and long term memory causes memory fragmentation.

Limited memory fragmentation is generally not an issue, when there is plenty of memory to spare, but for large model training running with limited memory, memory fragmentation leads to two issues, i) OOM due to lack of contiguous memory even when there is enough available memory, ii) poor efficiency as a result of the memory allocator spending significant time to search for a contiguous piece of memory to satisfy a memory request.

\name does memory defragmentation on-the-fly by pre-allocating contiguous memory chunks for activation checkpoints and gradients, and copying them over to the pre-allocated memory as they are produced. M$_D$ not only enables \name to train larger models with larger batch sizes, but also improves efficiency when training with limited memory.

\section{Communication Analysis of \name-DP}\label{sec:communication}
%{\color{green} [a figure on when/how communication is conducted could help this section, but optional for the time being.]}

%\name can enable training up to a  128 B   parameter model with 64-way data parallelism ($16 \times 128 / 64 = 32 GB$), compared to 2B parameter model ($16 \times 2 =32 GB$) without \name. This is 64x increase in the model size compared to what is currently possible (without model parallelism), and the size continues to grow with data parallelism. 

As \name boosts model size by removing memory redundancy, it is only natural to ask if we are trading communication volume for memory efficiency. In other words, what is the communication volume of \name-powered DP approach compared to a baseline DP approach?
The answer is in two parts: i) \name-DP incurs no additional communication using $P_{os}$ and $P_g$, while enabling up to 8x memory reduction,
%, resulting in a total memory footprint of $2\Psi + \frac{14\Psi}{N_d}$ 
ii) \name-DP incurs a maximum of $1.5$x communication when using $P_p$ in addition to $P_{os}$ and $P_{g}$, while further reducing the memory footprint by $N_d$ times.
%, resulting in a total memory foot print of $\frac{16\Psi}{N_d}$. 
We present the analysis in this section.  
We begin by first presenting a brief overview of the communication volume for standard DP. 

\subsection{Data Parallel Communication Volume}
During data parallel training, gradients across all data parallel processes are averaged at the end of the backward propagation before computing the updates for the next step. The averaging is performed using an all-reduce communication collective. For a large model size, the all-reduce communication is entirely communication bandwidth bound, and therefore, we limit our analysis to the total communication volume send to and from each data parallel process.

State-of-art implementation of all-reduce uses a two-step approach, where the first step is a reduce-scatter operation, which reduces different part of the data on different process. The next step is an all-gather operation where each process gathers the reduced data on all the process. The result of these two steps is an all-reduce.
Both reduce-scatter and all-gather are implemented using a pipelined approach, that results in a total data movement of $\Psi$ elements (for a data with $\Psi$ elements) for each. Therefore, the standard DP incurs 2$\Psi$ data movement during each training step.
%{\color{green} [Discuss and summarize the properties of \name corresponding to those mentioned in the overview.]}
 \begin{table*}
        \centering
        %\resizebox{\columnwidth}{!}{
            \begin{tabular}{|c|c||c|c|c|c||c|c|}
            \hline
            \multicolumn{1}{|c}{\multirow{2}{*}{MP}}&\multicolumn{1}{|c}{\multirow{2}{*}{GPUs}}  & \multicolumn{4}{||c|}{\bf Max Theoretical Model Size}&\multicolumn{2}{|c|}{\bf Measured Model Size} \\
            \hhline{~~------}
            &&Baseline&P$_{os}$&P$_{os+g}$&P$_{os+g+p}$&Baseline&\name-DP (P$_{os}$)\\
            \hline
            1&64&2B&\bf{7.6B}&14.4B&128B&1.3B&\bf{6.2B}\\
            \hline
            2&128&4B&\bf{15.2B}&28.8B&256B&2.5B&\bf{12.5B}\\
            \hline
            4&256&8B&\bf{30.4B}&57.6B&0.5T&5B&\bf{25B}\\
            \hline
            8&512&16B&\bf{60.8B}&115.2B&1T&\it{10B}&\bf{50B}\\
            \hline
            16&1024&32B&\bf{121.6B}&230.4B&\it{2T}&20B&\bf{100B}\\
            \hline
            \end{tabular}
        %}
        
     \caption{Maximum model size through memory analysis (left) and the measured model size when running with \nameos (right). The measured model size with $P_{os}$ matches the theoretical maximum, demonstrating that our memory analysis provides realistic upper bounds on model sizes.}
     \label{tab:largest-model}
 \end{table*}
\subsection{\name-DP Communication Volume}
\subsubsection{Communication Volume with $P_{os+g}$} 
With gradient partitioning, each process only stores the portion of the gradients, that is required to update its corresponding parameter partition. As such, instead of an all-reduce, \name only requires a scatter-reduce operation on the gradients, incurring communication volume of $\Psi$.  After each process updates the partition of the parameters that it is responsible for, an all-gather is performed to collect all the updated parameters from all the data parallel process. This also incurs a communication volume of $\Psi$. So the total communication volume per training step is $\Psi+\Psi= 2\Psi$, exactly the same as the baseline DP.

\subsubsection{Communication Volume with $P_{os+g+p}$ } 
After parameter partitioning, each data parallel process only stores the parameters that it updates. Therefore, during the forward propagation it needs to receives the parameters for all the other partitions. However, this can be pipelined to avoid the memory overhead. Before computing the forward propagation on the part of the model corresponding to a particular partition, the data parallel process responsible for that partition can broadcast the weights to all the data parallel processes. Once the forward propagation for that partition is done, the parameters can be discarded. The total communication volume is thus $\frac{\Psi \times N_d}{N_d} = \Psi$. In other words, we reschedule the parameter all-gather by spreading it across the entire forward propagation, and discarding the parameters once they have been used. Note however that this all-gather needs to happen once again for the backward propagation in the reverse order. 

The total communication volume is therefore the sum of the communication volumes incurred by these all-gathers in addition to the communication volume incurred by the reduce-scatter of the gradients. The total volume is therefore $3\Psi$ which is 1.5x compared to the baseline.   Both gradient and parameter partitioning leverage the insight that --- not all states of gradients and parameters are needed all the time --- to optimize memory by communicating the states judiciously. 

\begin{comment}
\subsection{Communication Latency}
Here we want to point out that while our optimizations impact communication latency, its impact to overall performance is likely small. Note that $P_g$ implements a reduce-scatter as a sequential series of reduce operations, and $P_p$ implements all-gather as a sequential series of broadcast operations.  Furthermore, $C_b$ can partition a large communication collective into multiple smaller ones to avoid memory overhead. Clearly, these sequential operations increases the communication latency. However, for large models with hundreds of billions of parameters, even with a large enough constant buffer size $C_B$, message sizes are large enough, that the communication time is bounded by the communication volume and communication bandwidth, and not by latency.
\end{comment}

\section{Communication Analysis of \name-R}
We compare the communication volume of partitioned activation checkpointing ($P_a$) in \name-R with baseline MP, and show that $P_a$ incurs a communication volume increase that is in general less than one tenth of the baseline MP. Furthermore, we analyze the communication overhead of $P_a$ in relation to DP communication volume to identify scenarios when $P_a$ improves efficiency by allowing for a larger batch size and reducing DP communication.  We leverage such analysis to decide if and when to apply  $P_a$ as well as $P_{a+cpu}$.  

Communication volume trade-off of partitioning activation checkpoints depends on the model size, checkpointing strategy and the MP strategy.  To share concrete insights, we perform the analysis in the context of transformer based models implemented using SOTA MP approach, Megatron-LM.

In Megatron-LM with activation checkpointing, each transformer block performs two all-reduce operations of size $batch \times seq\_length \times hidden\_dim$ in the forward propagation, two all-reduce for forward re-computation and two more in the backward propagation.  The total communication per block is $12 \times seq\_length \times hidden\_dim$ since communication volume of an all-reduce is $2 \times message\_size$. 

When \name-R partitions activation checkpoints, it requires an additional all-gather operation before the forward recomputation of the back-propagation on each activation checkpoint. In general, we checkpoint the input activation for each transformer block, requiring one all-gather per transformer block. The communication overhead $P_a$ is therefore $seq\_length * hidden\_dim$, since the communication volume of an all-gather is $message\_size$. Therefore, the total communication overhead of $P_a$ is less than $10\%$ of the original communication volume for model parallelism.

When MP is used in conjunction with DP, $P_a$ can be used to reduce the data-parallel communication volume by an order of magnitude at the expense of a $10\%$ increase in model-parallel communication volume, and significantly boost efficiency when data-parallel communication is a performance bottleneck. Notice that $P_a$ reduces the activation memory consumption by the MP degree allowing for a proportional increase in batch size. For large models, MP can be as large as 16 (\#GPUs on a DGX-2 node), allowing for up to 16x increase in the batch size. The communication volume of a data-parallel training is inversely proportional to the batch size. Therefore, an order of magnitude increase in batch size due to $P_a$ could result in an order-of-magnitude decrease in data-parallel communication volume. 

Finally if $P_{a+cpu}$ is applied, partitioned activation checkpoints are offloaded to CPU, reducing the activation memory requirement to nearly zero at the expense of 2x added data movement to and from CPU memory compared to $P_a$. In extreme cases where DP communication volume is the major bottleneck due to a small batch size even with $P_a$, $P_{a+cpu}$ can improve efficiency by increasing the batch size as long as the CPU data transfer overhead is less than the DP communication volume overhead, which is generally true for small batch sizes.  

Given model and hardware characteristics, we leverage the above analysis to decide if and when to apply $P_a$ and $P_{a+cpu}$.

\section{Step Towards 1 Trillion Parameters}
The largest published models today are in the range of 10 billion parameters, which are already challenging to train.  Getting to a trillion parameters,  3-orders of magnitude larger, will inevitably happen, but the road will be full of hurdles, surprises and innovations. While we do not claim knowing or addressing all of them, \name addresses one of the most fundamental challenges from a system perspective: the ability to fit a model of this scale on current hardware while allowing it to train with good system scalability.

\textbf{A Leap from State-of-Art}  
The largest model that the state-of-art framework, Megatron, can train with acceptable throughput is a 16 - 20B parameter model in a DGX-2 system. Scaling further by having model parallelism across multiple DGX nodes results in significant efficiency drop due to limited internode bandwidth.  

\name vastly increase the efficiently-runnable model size.  It enables the current generation of hardware to run significantly larger models without requiring fine-grained model parallelism to go across the node boundaries.
As demonstrated in Table \ref{tab:memory-consumption}, \name, with all optimizations turned on (P$_{os+g+p}$), could fit more than 1 \emph{Trillion} parameters on 1024 GPUs using DP only.  Alternatively, when combined with model parallelism (as shown in Table \ref{tab:largest-model}), \name could fit more than 1 \emph{Trillion} parameters on 1024 GPUs with 16-way model parallelism (within each DGX2 node) and 64-way data parallelism across nodes.  Running a model with a trillion parameters efficiently is no longer impossible!

\textbf{Compute Power Gap} Training a trillion parameter model end-to-end within an acceptable time range, however, could still require significant amount of compute power, which is lacking in today's AI clusters.

%Indeed, we can fit a model with over a trillion parameters using 16-way model parallelism, and 64-way data parallelism, and train it effectively since we are neither crossing the node boundary, or increasing the inter-node data-movement significantly (as discussed in Sec. ~\ref{communication-volume}). However, we would like to caution the reader, that training such a model in a reasonable time would still require significant amount of GPU resources. 

To understand the resource requirement, we present a brief comparison with Bert-Large. Bert-Large can be trained in $67$ minutes on a $1024$ GPU DGX-2H cluster \cite{nvidia-bert-training-with-gpus}. A 1 Trillion Parameter model can easily contain $3000$x (1 trillion / 330 million) more computation than a Bert-Large model for a data sample.  Even if we assume the same sequence length and the total number of samples required to train the model, training a 1T model would take 140 days, assuming the same hardware and similar computational efficiency.  In practice, both data samples and sequence length are likely to increase with the increased model size requiring over a year to train.  It would require an exa-flop system to train a 1T parameter model in a reasonable time. But when such compute capacity becomes available, we hope \name will provide the system technology to run the 1T models efficiently.

\begin{figure}
   \begin{minipage}[b]{0.55\columnwidth}
    \centering
       \includegraphics[width=\textwidth]{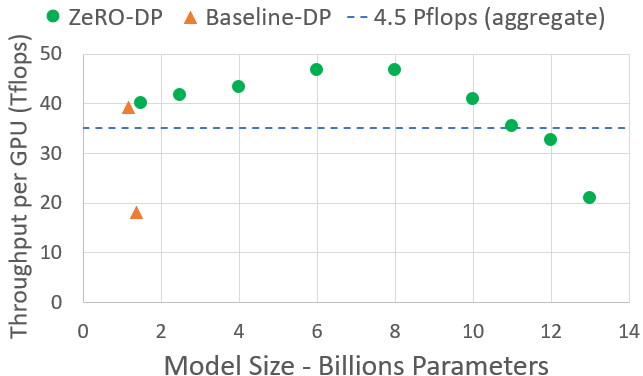}
        \caption{Max model throughput with \name-DP.} \label{fig:dp_tput}
   \vspace{0.04in}
   \end{minipage}
%   \vspace{-0.13in}
    \quad
   \begin{minipage}[b]{0.4\columnwidth}
    \centering
       \includegraphics[width=\textwidth]{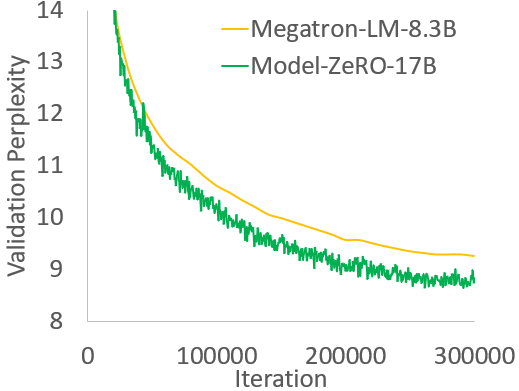}
        \caption{SOTA Turing-NLG enabled by \name.} \label{fig:turing_nlg_17B}
      \vspace{0.08in}
   \end{minipage}
\end{figure}
\section{Implementation and Evaluation}\label{sec:evaluation}

We focus our implementation on supporting efficient training of models with {$\sim$}100B parameters, which are an order-of-magnitude larger than the largest published models today (e.g., T5-11B~\cite{T5}) while trainable within a reasonable time frame on current hardware (e.g., with 1K V100 GPUs).  We implement and evaluate a subset of optimizations in \name --- $P_{os+g}$ in \name-DP plus ZeRO-R --- that allows us to achieve this goal. We will refer to this implementation as \name-100B.  Our results show that \name-100B can efficiently train models with up to 170B parameters, 8x bigger than SOTA, up to 10x faster and with improved usability. \name-100B powers Turing-NLG, the largest published model in the world with new SOTA accuracy.   

%Here, we demonstrate the performance of \name for models with up to 170B parameters. We show that \name-100B achieves better than perfect scalability in the regime that we tested. We show how \name democratizes large model training by providing data scientist the freedom to experiment without constraints, and making large model training feasible on lower end clusters. Furthermore, we discuss the impact of different optimization in \name-100B on model size, memory consumption and performance. We also present the \name-100B powered Model-\name, a 17.2B parameter transformer based model which is not only the SOTA for language models but is also the largest model in the world. We begin by presenting a brief discussion of our implementation and methodology.

\begin{table}
\centering
   \begin{minipage}[b]{0.30\columnwidth}
       \includegraphics[width=\textwidth]{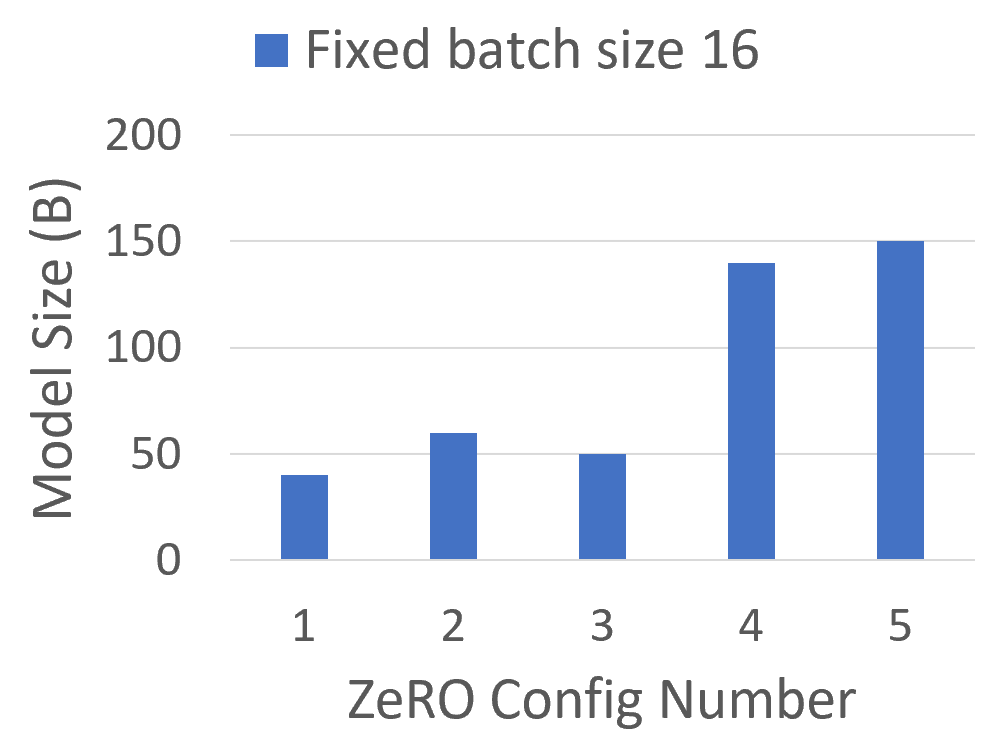}
        \captionof{figure}{Max model size \vspace{2pt}}. \label{fig:max-model-size}
   \end{minipage}    
   \quad
   \begin{minipage}[b]{0.30\columnwidth}
       \includegraphics[width=\textwidth]{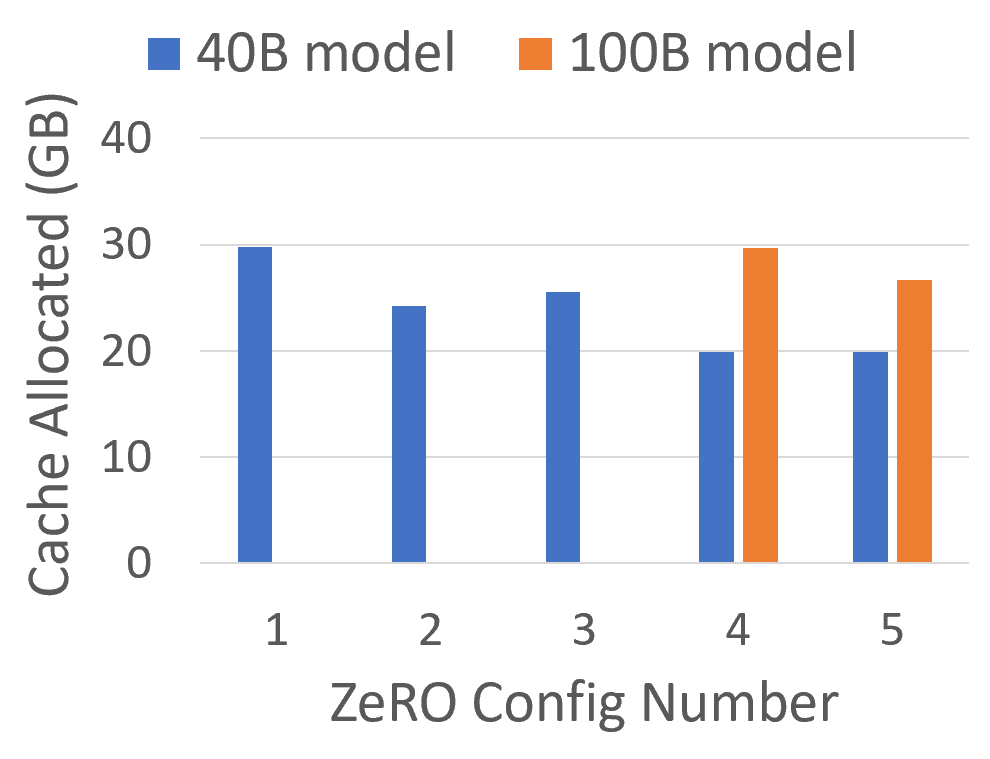}
       \captionof{figure}{Max cache allocated.} \label{fig:max-cached-memory}
   \end{minipage}    
   \quad
   \begin{minipage}[b]{0.30\columnwidth}
       \includegraphics[width=\textwidth]{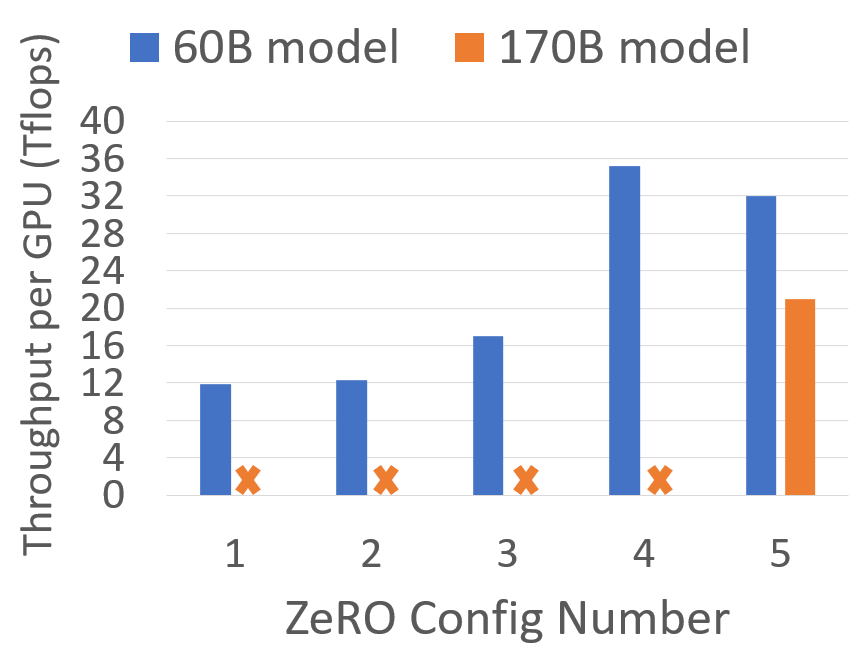}
        \captionof{figure}{\small{Throughput per GPU}.} \label{fig:max-performance}
   \end{minipage}
   \end{table}
  
\subsection{Implementation and Methodology}

%\paragraph{\name and MP} 

\paragraph{Implementation}
We implemented \name-100B in PyTorch including the full set of optimizations in $P_{os+g}$ and \name-R.  Its interface is compatible with any model implemented as an {\tt torch.nn.module}. Users can simply wrap their models using this interface and leverage \name-powered DP as they use classic DP. Users do not need to modify their model.  \name-powered DP can be combined with any form of MP including Megatron-LM.

\paragraph{Hardware}
We conducted our experiments on a cluster of 400 V100 GPUs ($25$ DGX-2 nodes) with 800 Gbps internode communication bandwidth.
%Details of our hardware environment are summarized in Table~\ref{tab:hardware_environment}. 

\paragraph{Baseline}
For experiments without MP, we use torch's distributed data parallel (DDP) as baseline. For experiments with MP, we use Megatron-LM because it is, to our knowledge, the state-of-art. We use the open-source version of Megatron-LM from NVIDIA~\footnote{https://github.com/nvidia/Megatron-LM} with a date of September 2019. The most recent Megatron-LM results report the ability to scale up to 16B parameter models using 32 DGX-2 nodes (total of 512 32GB V100 GPUs) \cite{megatronlm}. %Megatron-LM also comes with its own data-parallel implementation.

\paragraph{\name}
Experiments without MP, use the \name-powered DP implementation in \name-100B. 
%As discussed in Sec.~\ref{sec:intro}, \name can be combined with MP to achieve a multiplicative effect in memory reduction. We use \name in conjunction with MP to efficiently train very large model sizes. 
Experiments with MP, combine \name-powered DP with MP of Megatron-LM.

\paragraph{Model Configurations}
The models presented in this section are GPT-2 \cite{gpt-2} like transformer based models. We vary the hidden dimension and the number of layers to obtain models with different number of parameters. Table~\ref{tab:model-configuration} shows the configuration parameters used in our experiments with additional details in AE Appendix.

\subsection{Speed and Model Size}
\name-100B efficiently run models with up to 170B parameters on 400 GPUs, more than 8x bigger than Megatron-LM. Figure~\ref{fig:billion_parameter_speedup} shows throughput per GPU for varying model sizes using \name-100B with MP versus using Megatron MP alone. \name-100B achieves a sustained throughput of 15 PetaFlops (over 30\% of the peak) on average for models with 8B to 100B parameters. In comparison, the baseline MP performance degrades quickly with the increase in model size: MP incurs high communication volume between GPUs, and going beyond a single node to fit larger models causes a communication bandwidth drop from 300GB/sec per link (NVSwitch) to 12.5 GB/sec per link (Infiniband EDR), resulting in a significant performance drop. \name-100B achieves up to 10x speedup over baseline, significantly outperforming on large models.

For \name-100B, the slight reduction in performance beyond 100B is due to lack of enough memory to run larger batch sizes. We expect the performance to improve as we increase the number of GPUs due to super-linear speedup of \name-100B as we discuss next. 
\begin{table}\centering
         \begin{tabular}{|c|c|c|}
         \hline
         & \name-DP & \name-R \\
         \hline
         1 & P$_{os}$&C$_{B}$+M$_{D}$\\
         \hline
         2 & P$_{os}$&C$_{B}$+M$_{D}$+P$_{a}$\\
         \hline
         3 & P$_{os+g}$&C$_{B}$+M$_{D}$\\
         \hline
         4 & P$_{os+g}$&C$_{B}$+M$_{D}+$P$_{a}$\\
         \hline
         5 & P$_{os+g}$&C$_{B}$+M$_{D}$+P$_{a+cpu}$\\
         \hline
         \end{tabular}
     \caption{\name configurations}\label{tab:Opt-table}

\end{table}
\subsection{Super-Linear Scalability}
\name-100B demonstrates super-linear scalability for very large model sizes. Figure~\ref{fig:hyperscale_60B} shows scalability results for a 60B parameter model going from 64 to 400 GPUs and we expect this trend to continue further for more GPUs. $P_{os+g}$ reduces per GPU memory consumption of \name-100B with increase in DP degree, allowing \name-100B to fit larger batch sizes per GPU\footnote{Increasing batch size too much can lead to poor convergence, but for these large models, we are still in a regime where batch size is small enough even with 1K GPU and it does not affect convergence rate}, which in turn improves throughput as a result of increasing arithmetic intensity. 
%We realize that these results do not represent strong or weak scaling in the traditional sense. However, in practical terms, this scaling directly corresponds to the end-to-end wall-clock time which is the most meaningful metric to evaluate scalability for DL workloads.

% \begin{table}[]
% \begin{tabular}{|l|l|c|l|l|c|}
% \hline
% \multicolumn{3}{|c|}{Figure ~\ref{fig:billion_parameter_speedup}} & \multicolumn{3}{c|}{Figures ~\ref{fig:hyperscale_60B}, \ref{fig:dp_tput}} \\ \hline
% \multicolumn{1}{|c|}{} & \multicolumn{1}{c|}{Layers} & HD & \multicolumn{1}{c|}{} & \multicolumn{1}{c|}{Layers} & HD \\ \hline
% 1.5B & 48 & 1600 & 1.16B--2.5B & 24,34,54 & 1920 \\ \hline
% 8B & 72 & 3072 & 4B & 64 & 2304 \\ \hline
% 20B & 98 & 4096 & 6B-8B & 52,72 & 3072 \\ \hline
% 40B--60B & 88, 132 & 6144 & 10B--13B & 50,54,58,62 & 4096 \\ \hline
% 80B--170B & \begin{tabular}[c]{@{}l@{}}100,125,150,\\ 175,212\end{tabular} & 8192 & 60B & 75 & 8192 \\ \hline
% \end{tabular}
% \caption{Configurations for different model sizes, number of layers, and hidden dimensions (HD) across Figures~\ref{fig:billion_parameter_speedup}, \ref{fig:hyperscale_60B}, \ref{fig:dp_tput}.} \label{tab:model-configuration}
% \vspace{-0.10in}
% \end{table}

\subsection{Democratizing Large Model Training}
Using MP and PP is challenging for many data scientists, which is a well-known hurdle to train large models.  \name does not require any changes to the model itself and it can be used as simple as baseline DP while delivering significantly boosted model size and speed.  Fig.~\ref{fig:dp_tput} shows that \name-100B can train models with up to 13B parameters without MP on 128 GPUs, achieving throughput over 40\,TFlops per GPU on average. In comparison, without \name, the largest trainable model with DP alone has 1.4B parameters with throughput less than 20\,TFlops per GPU. 
Furthermore, in the absence of the communication overhead from MP, these models can be trained with lower-end compute nodes without very fast intra-node interconnect such as NVLINK or NVSwitch, which is required to achieve good efficiency with MP.  
%In fact, \name-100B enables a 10B model to train faster than the fastest model using DP alone, which was a 1.2B parameter model running at 39 TFlops per GPU based on our experiments. 

%With \name-100B, the largest models in literature, the T5 11B \cite{T5} and Megatron 8.3B \cite{megatronlm} can be efficiently trained without any form of MP or PP. This allows data scientist can experiment freely with large model sizes because unlike MP or PP, \name does not require any changes to the model itself. 

\subsection{Memory and Performance Analysis}
We look into the benefits and impact of different optimizations on maximum model size, memory consumption and performance. These optimizations are referred to as Config 1 to 5 (C1-C5) in Table.~\ref{tab:Opt-table}.

%present insights into the optimizations in \name-100B by demonstrating the impact of \name-DP (namely $P_{os}$ and $P_{os+g}$), \name-R and \name-R with CPU off-loading($P_{a+cpu})$ on maximum model size, memory consumption and performance. There optimizations are referred to as Config 1 though 5 (C1-C5) as shown in Table.~\ref{tab:Opt-table}.
\paragraph{Maximum Model Size}
Figure~\ref{fig:max-model-size} shows the largest trainable model by enabling different \name optimizations for a fixed batch size and MP of 16. The model size increase from 40B to 60B when trained with C1 vs C2  due to a 16x (MP degree) reduction in activation memory from using $P_a$, while the jump to 140B using C4 is from enabling $P_{os+g}$ which halves the memory requirement by the model states compared to $P_{os}$ in C2. The increase to 150B using C5 is solely due to further reduction in activation memory from offloading the partitioned activation checkpoints to the CPU memory.
\paragraph{Max Cached Memory}
Figure~\ref{fig:max-cached-memory} shows the maximum memory cached by PyTorch during each training iteration for a 40B and a 100B parameter model. The decrease of the cached memory size is as expected from C1 to C2. The difference in memory consumption between C2 and C3 depends on the size of the model states in comparison to the activation memory, and can increase when activation memory is larger, or decrease when the model states are larger. It is note worthy that the cached memory does not decrease from C4 to C5 for 40B but it does for 100B. This is simply because the activation memory for 100B is much larger for the decrease to be noticeable. This makes $P_{a+cpu}$ a valuable tool to fit a larger batch size when we get to very large models. In Figure~\ref{fig:max-performance}, $P_{a+cpu}$ is needed for 170B model to execute without running out of memory.

\begin{table}[]
\centering
\begin{tabular}{|l|l|l|l|l|l|}
\hline
\multicolumn{3}{|c|}{Figure ~\ref{fig:billion_parameter_speedup}} & \multicolumn{3}{c|}{Figures ~\ref{fig:hyperscale_60B}, \ref{fig:dp_tput}} \\ \hline
\multicolumn{1}{|c|}{} & \multicolumn{1}{c|}{Layers} & \multicolumn{1}{c|}{HD} & \multicolumn{1}{c|}{} & \multicolumn{1}{c|}{Layers} & \multicolumn{1}{c|}{HD} \\ \hline
1.5B      & 48          & 1600 & 1.16B-2.5B  & 24,34,54    & 1920 \\ \hline
8B        & 72          & 3072 & 4B          & 64          & 2304 \\ \hline
40B-60B   & 88,132      & 4096 & 6B-8B       & 52,72       & 3072 \\ \hline
80B-170B  & 100,125,150 & 8192 & 10B-13B     & 50,54,58,62 & 4096 \\ \hline
140B-170B & 175,212     & 8192 & 60B         & 75          & 8192 \\ \hline
\end{tabular}
\caption{Configurations for different model sizes, number of layers, and hidden dimensions (HD) across Figures~\ref{fig:billion_parameter_speedup}, \ref{fig:hyperscale_60B}, \ref{fig:dp_tput}.} \label{tab:model-configuration}
\end{table}

% \begin{table}
% \centering
% \begin{tabular}{|l|l|c|l|l|c|}
% \hline
% \multicolumn{3}{|c|}{Figure ~\ref{fig:billion_parameter_speedup}} & \multicolumn{3}{c|}{Figures ~\ref{fig:hyperscale_60B}, \ref{fig:dp_tput}} \\ \hline
% \multicolumn{1}{|c|}{} & \multicolumn{1}{c|}{Layers} & HD & \multicolumn{1}{c|}{} & \multicolumn{1}{c|}{Layers} & HD \\ \hline
% 1.5B & 48 & 1600 & 1.16B--2.5B & 24,34,54 & 1920 \\ \hline
% 8B & 72 & 3072 & 4B & 64 & 2304 \\ \hline
% 20B & 98 & 4096 & 6B-8B & 52,72 & 3072 \\ \hline
% 40B--60B & 88, 132 & 6144 & 10B--13B & 50,54,58,62 & 4096 \\ \hline
% 80B--170B & \begin{tabular}[c]{@{}l@{}}100,125,150,\\ 175,212\end{tabular} & 8192 & 60B & 75 & 8192 \\ \hline
% \end{tabular}
% \caption{Configurations for different model sizes, number of layers, and hidden dimensions (HD) across Figures~\ref{fig:billion_parameter_speedup}, \ref{fig:hyperscale_60B}, \ref{fig:dp_tput}.} \label{tab:model-configuration}
% \vspace{-0.095in}
% \end{table}

\paragraph{Max Achievable Performance}
Figure~\ref{fig:max-performance} shows the best achievable performance for different set of optimizations. Notice that performance improvement corresponds to decrease in memory consumption between the optimizations. As mentioned earlier, lower memory consumption allows for larger batch size which improves performance. The only caveat is the performance drop between C4 and C5 for 60B parameter model. Despite lower memory consumption, C5 incurs activation movement to and from the CPU, this will result in worse performance in most cases, except for a few where the model is so large that the model simply cannot run without C5 or the batch size that can run without C5 is very small (such as model with 170B parameters in Figure~\ref{fig:max-performance}). During training, $P_{a+cpu}$ is turned on only when it is beneficial.
\subsection{Turing-NLG, the SOTA language model with 17B parameters}
As of May 12th, 2020, Turing-NLG is the largest model in the world with over 17B parameters. It achieved the new SOTA for language models with Webtext-103 perplexity of 10.21. Turing-NLG was trained end-to-end using \name-100B and Fig.~\ref{fig:turing_nlg_17B} shows the validation perplexity over 300K iterations compared to previous SOTA, Megatron-LM 8.3B parameter model. \name-100B achieves a sustained throughput of 41.4\,TFlops/GPU for this model.

\section{Concluding Remarks}
From a HPC and system perspective, we believe that \name represents a revolutionary transformation in the large model training landscape. While our implementation, \name-100B, enables 8x increase in model sizes, over 10x in throughput improvement, achieves super-linear speedups on modern GPU clusters, and trains the largest model in the world, it is still just a tip of the iceberg. \name in its entirety has the potential to increase the model size by yet another order of magnitude, enabling the training of trillion parameter models of the future. 

Perhaps, what we feel most optimistic about \name is that it imposes no hurdles on the data scientists. Unlike existing approaches such as MP and PP, no model refactoring is necessary, and it is as easy to use as standard DP, making \name a prime candidate for future investigations on large model training. Through open sourcing and community feedback, we plan to make \name fully accessible to the DL community to catalyze the evolution and democratization of large model training at scale.  
\section*{Acknowledgement}

We thank Junhua Wang for his valuable support and advice. We thank Minjia
Zhang, Elton Zheng, Shaden Smith, Reza Yazdani Aminabadi, Arash Ashari, and
Niranjan Uma Naresh for their great feedback and help on evaluating the work.
We thank Brandon Norick, Corby Rossett, Gopi Kumar, Jack Zhang, Jing Zhao,
Payal Bajaj, Rangan Majumder, Saksham Singhal, Saurabh Tiwary, and Xia Song for
many helpful discussions and suggestions.

\bibliographystyle{unsrt}
\bibliography{references}

%TODO: remove below for submission
\newpage
\appendix
% \subsection{Experiments}
% In this appendix we outline all of the model parameters used for each of the corresponding figures in our paper.
% Experiment details for Figure 2 in the paper can be seen in Tables~\ref{tab:fig2-tab}, \ref{tab:fig3-tab}, \ref{tab:fig7-tab}.

\begin{table}[t!]
\scriptsize
\begin{tabular}{|l|l|l|l|l|l|l|l|l|}
\hline
\multicolumn{9}{|c|}{Figure 2} \\ \hline
Model size & ZeRO/Baseline & Number of GPUs & MP & Layers & Hidden size & Attention head & Batch size & Total batch size \\ \hline
1.5B & ZeRO & 400 & 1 & 48 & 1600 & 16 & 24 & 9600 \\ \hline
1.5B & Baseline & 400 & 2 & 48 & 1600 & 16 & 16 & 3200 \\ \hline
8B & ZeRO & 400 & 4 & 72 & 3072 & 24 & 64 & 6400 \\ \hline
8B & Baseline & 400 & 8 & 72 & 3072 & 24 & 8 & 400 \\ \hline
40B & ZeRO & 400 & 4 & 88 & 6144 & 32 & 12 & 1200 \\ \hline
40B & Baseline & 384 & 32 & 88 & 6144 & 64 & 4 & 48 \\ \hline
60B & ZeRO & 400 & 16 & 132 & 6144 & 32 & 64 & 1600 \\ \hline
60B & Baseline & 384 & 64 & 132 & 6144 & 64 & 4 & 24 \\ \hline
80B & ZeRO & 400 & 16 & 100 & 8192 & 64 & 32 & 800 \\ \hline
80B & Baseline & 384 & 128 & 100 & 8192 & 128 & 4 & 12 \\ \hline
100B & ZeRO & 400 & 16 & 125 & 8192 & 64 & 32 & 800 \\ \hline
100B & Baseline & 384 & 128 & 125 & 8192 & 128 & 2 & 6 \\ \hline
120B & ZeRO & 400 & 16 & 150 & 8192 & 64 & 24 & 600 \\ \hline
120B & Baseline & 384 & 128 & 150 & 8192 & 128 & 2 & 6 \\ \hline
140B & ZeRO & 400 & 16 & 175 & 8192 & 64 & 16 & 400 \\ \hline
140B & Baseline & 384 & 128 & 175 & 8192 & 128 & 2 & 6 \\ \hline
170B & ZeRO & 400 & 16 & 212 & 8192 & 64 & 12 & 300 \\ \hline
170B & Baseline & 256 & 256 & 212 & 8192 & 256 & 2 & 2 \\ \hline
\end{tabular}
\caption{Model configurations for Figure 2 related to ZeRO throughput compared with baseline.} \label{tab:fig2-tab}
\end{table}

\begin{table}[t!]
\scriptsize
\begin{tabular}{|l|l|l|l|l|l|l|l|l|}
\hline
\multicolumn{9}{|c|}{Figure 3} \\ \hline
Model size & ZeRO/Baseline & Number of GPUs & MP & Layers & Hidden size & Attention head & Batch size & Total batch size \\ \hline
60B & ZeRO & 64 & 16 & 75 & 8192 & 32 & 16 & 64 \\ \hline
60B & ZeRO & 128 & 16 & 75 & 8192 & 32 & 48 & 384 \\ \hline
60B & ZeRO & 256 & 16 & 75 & 8192 & 32 & 48 & 768 \\ \hline
60B & ZeRO & 400 & 16 & 75 & 8192 & 32 & 64 & 1600 \\ \hline
\end{tabular}
\caption{Model configurations for Figure 3 related to superlinear scalability.} \label{tab:fig3-tab}
\end{table}

\begin{table}[t!]
\scriptsize
\begin{tabular}{|l|l|l|l|l|l|l|l|l|}
\hline
\multicolumn{9}{|c|}{Figure 4} \\ \hline
Model size & ZeRO/Baseline & Number of GPUs & MP & Layers & Hidden size & Attention head & Batch size & Total batch size \\ \hline
40B & ZeRO & 400 & 16 & 50 & 8192 & 32 & 16 & 400 \\ \hline
60B & ZeRO & 400 & 16 & 132 & 6144 & 64 & 16 & 400 \\ \hline
140B & ZeRO & 400 & 16 & 175 & 8192 & 64 & 16 & 400 \\ \hline
150B & ZeRO & 400 & 16 & 187 & 8192 & 64 & 16 & 400 \\ \hline
50B & ZeRO & 400 & 16 & 62 & 8192 & 32 & 16 & 400 \\ \hline
\end{tabular}
\caption{Model configurations for Figure 4 related to max model size with different ZeRO configurations.} \label{tab:fig4-tab}
\end{table}

\begin{table}[t!]
\scriptsize
\begin{tabular}{|l|l|l|l|l|l|l|l|l|}
\hline
\multicolumn{9}{|c|}{Figure 5} \\ \hline
Model size & ZeRO/Baseline & Number of GPUs & MP & Layers & Hidden size & Attention head & Batch size & Total batch size \\ \hline
40B & ZeRO & 400 & 16 & 50 & 8192 & 32 & 16 & 400 \\ \hline
40B & ZeRO & 400 & 16 & 50 & 8192 & 32 & 16 & 400 \\ \hline
40B & ZeRO & 400 & 16 & 50 & 8192 & 32 & 16 & 400 \\ \hline
40B & ZeRO & 400 & 16 & 50 & 8192 & 32 & 16 & 400 \\ \hline
40B & ZeRO & 400 & 16 & 50 & 8192 & 32 & 16 & 400 \\ \hline
100B & ZeRO & 400 & 16 & 125 & 8192 & 64 & 32 & 800 \\ \hline
100B & ZeRO & 400 & 16 & 125 & 8192 & 64 & 32 & 800 \\ \hline
\end{tabular}
\caption{Model configurations for Figure 5 related to memory allocated with different ZeRO configurations.}
\end{table}

\begin{table}[t!]
\scriptsize
\begin{tabular}{|l|l|l|l|l|l|l|l|l|}
\hline
\multicolumn{9}{|c|}{Figure 6} \\ \hline
Model size & ZeRO/Baseline & Number of GPUs & MP & Layers & Hidden size & Attention head & Batch size & Total batch size \\ \hline
60B & ZeRO & 128 & 16 & 75 & 8192 & 64 & 2 & 16 \\ \hline
60B & ZeRO & 128 & 16 & 75 & 8192 & 64 & 4 & 32 \\ \hline
60B & ZeRO & 128 & 16 & 75 & 8192 & 64 & 32 & 256 \\ \hline
60B & ZeRO & 128 & 16 & 75 & 8192 & 64 & 32 & 256 \\ \hline
60B & ZeRO & 128 & 16 & 75 & 8192 & 64 & 8 & 64 \\ \hline
170B & ZeRO & 400 & 16 & 212 & 8192 & 64 & 12 & 300 \\ \hline
\end{tabular}
\caption{Model configurations for Figure 6 related to throughput with different ZeRO configurations.}
\end{table}
.
\newpage
These tables contain all the model configurations and batch sizes used for the experiments presented in the paper. In Figure 2, notice that the total number of GPUs for some baseline experiment is 384 or 256 compared to 400 for ZeRO. This is because the total number of GPUs must be a product of the number of MP, and we only had access to a total of 400 GPUs. There exist a handful of additional constraints in model configuration values, such as hidden size must be divisible by attention heads, hidden size divisible by MP, and attention heads divisible by MP. For baseline we used the lowest number of GPUs that was a power of 2 that would fit the model. So for example, for 170B parameter model this was 256 for the baseline. Since we only had 400 GPUs, we could only run baseline with 256 GPUs. 

We do want to point out that this gives the baseline an advantage over ZeRO because fewer GPUs means better communication throughput for the baseline. For example, in case of the 170B parameter model, DP=1 for the baseline so it in fact incurs no communication for DP. The results presented in this paper are despite this advantage for the baseline.

Also, we want to point out that we are comparing the performance per GPU, not the aggregate performance, and therefore the results are still apples-to-apples while giving a slight advantage to the baseline.

\begin{table}[]
\scriptsize
\begin{tabular}{|l|l|l|l|l|l|l|l|l|}    
\hline
\multicolumn{9}{|c|}{Figure 7} \\ \hline
Model size & ZeRO/Baseline & Number of GPUs & MP & Layers & Hidden size & Attention head & Batch size & Total batch size \\ \hline
1.5B & ZeRO & 128 & 1 & 34 & 1920 & 16 & 24 & 3072 \\ \hline
2.5B & ZeRO & 128 & 1 & 54 & 1920 & 16 & 24 & 3072 \\ \hline
4B & ZeRO & 128 & 1 & 64 & 2304 & 24 & 16 & 2048 \\ \hline
6B & ZeRO & 128 & 1 & 52 & 3072 & 24 & 12 & 1536 \\ \hline
8B & ZeRO & 128 & 1 & 72 & 3072 & 24 & 8 & 1024 \\ \hline
10B & ZeRO & 128 & 1 & 50 & 4096 & 32 & 6 & 768 \\ \hline
11B & ZeRO & 128 & 1 & 54 & 4096 & 32 & 4 & 512 \\ \hline
12B & ZeRO & 128 & 1 & 58 & 4096 & 32 & 4 & 512 \\ \hline
13B & ZeRO & 128 & 1 & 62 & 4096 & 32 & 2 & 256 \\ \hline
1p16B & Baseline & 128 & 1 & 24 & 1920 & 16 & 8 & 1024 \\ \hline
1p38B & Baseline & 128 & 1 & 40 & 1536 & 16 & 1 & 128 \\ \hline
\end{tabular}
\caption{Model configurations for Figure 7 related to evaluating maximum model sizes vs throughput while using only data-parallelism.} \label{tab:fig7-tab}
\end{table}

%TODO: remove above for submission

\end{document}